\documentclass[sigconf]{acmart}

\usepackage{graphicx}
\usepackage{subcaption}
\usepackage{enumitem}
\usepackage{amsmath}
\usepackage{hyperref}
\usepackage{url}
\AtBeginDocument{%
  }
\usepackage[utf8]{inputenc}
\usepackage{xcolor} 

\usepackage{float}

\usepackage{lipsum}
\copyrightyear{2025}
\acmYear{2025}
\setcopyright{rightsretained}
\acmConference[WiSec 2025]{18th ACM Conference on Security and Privacy in Wireless and Mobile Networks}{June 30-July 3, 2025}{Arlington, VA, USA}
\acmBooktitle{18th ACM Conference on Security and Privacy in Wireless and Mobile Networks (WiSec 2025), June 30-July 3, 2025, Arlington, VA, USA}
\acmDOI{10.1145/3734477.3736147}
\acmISBN{979-8-4007-1530-3/2025/06}
\begin{document}

\title{POSTER: A Multi-Signal Model for Detecting Evasive Smishing}

\author{Shaghayegh Hosseinpour}
\affiliation{%
  \institution{George Mason University}
  \city{Fairfax}
  \state{Virginia}
  \country{USA}
}
\email{shosse2@gmu.edu}
\author{Sanchari Das}
\affiliation{%
  \institution{George Mason University}
  \city{Fairfax}
  \state{Virginia}
  \country{USA}
}
\email{sdas35@gmu.edu}

\renewcommand{\shortauthors}{Shaghayegh Hosseinpour and Sanchari Das}

\begin{abstract}
Smishing, or SMS-based phishing, poses an increasing threat to mobile users by mimicking legitimate communications through culturally adapted, concise, and deceptive messages, which can result in the loss of sensitive data or financial resources. In such, we present a multi-channel smishing detection model that combines country-specific semantic tagging, structural pattern tagging, character-level stylistic cues, and contextual phrase embeddings. We curated and relabeled over $84,000$ messages across five datasets, including $24,086$ smishing samples. Our unified architecture achieves 97.89\% accuracy, an F1 score of $0.963$, and an AUC of $99.73\%$, outperforming single-stream models by capturing diverse linguistic and structural cues. This work demonstrates the effectiveness of multi-signal learning in robust and region-aware phishing.
\end{abstract}



\maketitle

\section{Introduction}
Short Message Service (SMS) is widely used for its accessibility and low cost, but its widespread use makes it a common target for smishing, a form of phishing via SMS that impersonates trusted entities to steal sensitive information~\cite{tally2023mid,gopavaram2021cross}. Unlike traditional spam, smishing messages are often short, informal, and regionally adapted, mimicking banks, delivery services, or government agencies. These messages frequently embed URLs, phone numbers, or emotionally manipulative language to bypass user suspicion and automated filters~\cite{blancaflor2024unmasking}. Despite recent advancements in spam and phishing detection~\cite{noah2022phishercop}, many models focus on only one aspect, such as shallow lexical features or token-level representations, and fail to capture the full range of nuanced, culturally adaptive elements of smishing. Attackers increasingly use informal syntax, deceptive formatting, and benign-sounding phrases to exploit user trust and evade detection. This highlights the need for more holistic detection strategies that integrate semantic, structural, stylistic and phrase-sensitive cues to identify malicious intent in highly variable and low-context text environments like SMS. In response, we explore the combination of country-specific tagging, textual pattern tagging, character-level features, and contextual phrase embeddings to better understand the linguistic and behavioral traits of smishing. Using a large-scale, relabeled corpus of over $84,000$ SMS messages drawn from five public datasets, we conduct an in-depth analysis and demonstrate how multi-faceted feature extraction can enable more robust and interpretable detection.

\section{Background}
Prior research works in smishing detection works have explored both traditional and deep learning approaches~\cite{kyaw2024systematic}. Early models, such as Naive Bayes and SVMs, used lexical features like message length and keyword frequency~\cite{de2024detection}, but often failed against obfuscated or region-specific messages. Deep learning models including CNNs~\cite{roumeliotis2024next}, LSTMs~\cite{sri2024improved}, and transformers improved accuracy~\cite{jamal2024improved}, yet struggled with informal syntax, character-level manipulation, and reused benign phrases. Recent efforts addressed regional adaptation by incorporating named entities and local context~\cite{gupta2024detection}, while lightweight models targeted on-device deployment. However, most existing systems still rely on single-feature pipelines~\cite{rustam2024deteksi} and overlook other cues essential for robust smishing detection.

\section{Methodology}
We built our detection system by fusing four types of features, concatenating them into a unified feature representation for each message. We combined data from ExAIS\_SMS \cite{abayomi2022deep}, Smishtank \cite{timko2024smishing}, Real-time SMS dataset \cite{mishra2022implementation}, SMS Spam Collection \cite{hidalgo2012validity}, and the Super dataset \cite{salman2025spallm}. Messages in these datasets were labeled as spam or non-spam, with spam messages further classified as smishing if they contained frequent smishing words \cite{shaghayegh2025smishing}.

\textbf{1. Country-specific Feature Extraction:}
We applied named entity recognition (NER) to extract geopolitical regions, organizations, and currencies, replacing them with abstract labels (\texttt{[GPE]}, \texttt{[ORG]}, \texttt{[MONEY]}) and appending the detected country name to each message. This preserved context while promoting generalization. We then vectorized the tagged messages using TF-IDF and trained a Random Forest classifier on the features.

\textbf{2. Textual Feature Tagging:}
We used regular expressions to tag structural elements like URLs, emails, and phone numbers with placeholders (e.g., \texttt{[URL]}, \texttt{[PHONE]}, \texttt{[EMAIL]}). This helped the model focus on intent patterns. We then applied TF-IDF and trained a Random Forest classifier.

\textbf{3. Character-Level Modeling:}
To capture informal syntax, emojis, and obfuscation tactics, we implemented a CharCNN. We converted each message into a fixed-length sequence of character indices, embedded them into dense vectors, and passed them through a one-dimensional convolutional layer followed by max pooling and dense layers. This channel captured stylistic manipulations often used to evade word-based filters.

\textbf{4. Contextual Phrase Embeddings:}  
We tagged curated benign and suspicious phrases with labels (e.g., \texttt{[legitimate\_like]}, \texttt{[smishing\_like]}), embedded the messages using DistilBERT, and extracted the [CLS] token. A CNN-based classifier captured phrase semantics. We then reduced each feature stream with Truncated SVD, concatenated the results, and trained a multi-layer perceptron with dropout and attention using binary cross-entropy and the Adam optimizer on an 80/20 train-test split.

\section{Results}
We evaluated each feature channel independently, then assessed performance after fusing all streams to understand individual contributions and the value of multi-channel integration.

\textbf{Semantic Stream:} Using NER to tag countries, organizations, and currencies, this stream achieved $96.34\%$ accuracy. It was effective at detecting regional scam patterns—e.g., ``India" appeared in both smishing and legitimate messages, ``US" mostly in non-smishing, while ``UK" appeared predominantly in smishing.

\textbf{Structural Stream:} Tagging URLs, email addresses, and phone numbers captured key phishing indicators. This stream slightly outperformed the semantic stream with 96.46\% accuracy.

\textbf{Character-Level Stream:} Our CharCNN achieved the highest individual accuracy at $96.62\%$, capturing stylized writing, emojis, and character-level obfuscation missed by token-based models.

\textbf{Contextual Phrase Stream:} Though it performed lowest in isolation ($88.29\%$), it helped the fused model by identifying repeated phishing phrases and tonal cues.

\textbf{Full Model Performance:} Combining all four channels yielded $97.89\%$ accuracy, $0.963$ F1 score, and $99.73\%$ AUC. Each stream added value; ablation studies showed notable performance drops when any stream was removed, especially the character and structural channels. Table~\ref{tab:results} summarizes the performance of each stream and the final model.

\begin{table}[h]
\small 
\setlength{\tabcolsep}{6pt}
\centering
\caption{Performance of Feature Streams and Proposed Combined Model}
\begin{tabular}{lccc}
\toprule
\textbf{Stream} & \textbf{Accuracy} & \textbf{F1} & \textbf{AUC} \\
\midrule
Semantic (NER) & 96.34\% & 0.934 & 95.2\% \\
Structural (Lexical) & 96.46\% & 0.936 & 95.3\% \\
Char-Level (CharCNN) & 96.62\% & 0.937 & 99.3\% \\
Contextual (DistilBERT) & 88.29\% & 0.789 & 94.0\% \\
\textbf{Combined Model} & \textbf{97.89\%} & \textbf{0.963} & \textbf{99.73\%} \\
\bottomrule
\end{tabular}
\label{tab:results}
\end{table}

\section{Conclusion}
Smishing is an effective form of social engineering that takes advantage of the informal and low-context nature of SMS. To address this threat, we developed a multi-channel detection model that integrates structural, semantic, stylistic, and contextual signals to identify both clear and subtle smishing patterns. Our model achieved $97.89\%$ accuracy, an F1 score of $0.963$, and an AUC of $99.73\%$, outperforming models that rely on a single type of feature. These results highlight the limitations of shallow and isolated approaches. The model's modular design allows flexible deployment across platforms, with lightweight components such as structural tagging and character-level analysis running on mobile devices, and semantic processing handled in the cloud. In future, we plan to explore region-specific adaptation, real-time detection, and support for multiple languages to improve smishing defenses further.

\bibliographystyle{ACM-Reference-Format}
\bibliography{sample-base.bib}

\end{document}